\documentclass[conference]{IEEEtran}
\IEEEoverridecommandlockouts
\usepackage{cite}
\usepackage{amsmath,amssymb,amsfonts}
\usepackage{algorithmic}
\usepackage{graphicx}
\usepackage{textcomp}
\usepackage{xcolor}
\usepackage{tabularx}
\usepackage{xcolor,colortbl}
\usepackage{siunitx} 
\usepackage{algorithm2e}
\usepackage{pgfplots}
\usepackage{tikz}
\usepackage{threeparttablex}
\usepackage{float}

\definecolor{darkgrey}{rgb}{0.5,0.5,0.5}

\makeatletter
\newcommand{\linebreakand}{%
  \end{@IEEEauthorhalign}
  \hfill\mbox{}\par
  \mbox{}\hfill\begin{@IEEEauthorhalign}
}
\makeatother

\def\BibTeX{{\rm B\kern-.05em{\sc i\kern-.025em b}\kern-.08em
    T\kern-.1667em\lower.7ex\hbox{E}\kern-.125emX}}
\begin{document}

\title{Low-latency Visual Previews of Large Synchrotron
Micro-CT Datasets\\
}
\author{\IEEEauthorblockN{Nicholas Tan Jerome}
\IEEEauthorblockA{\textit{Karlsruhe Institute of Technology (KIT)}\\
\textit{Institute for Data Processing and Electronics} \\
Eggenstein-Leopoldshafen, Germany \\
nicholas.tanjerome@kit.edu}
\and
\IEEEauthorblockN{Suren Chilingaryan}
\IEEEauthorblockA{\textit{Karlsruhe Institute of Technology (KIT)}\\
\textit{Institute for Data Processing and Electronics} \\
Eggenstein-Leopoldshafen, Germany \\
suren.chilingaryan@kit.edu}
 \linebreakand
\IEEEauthorblockN{Thomas van de Kamp}
\IEEEauthorblockA{\textit{Karlsruhe Institute of Technology (KIT)}\\
\textit{Institute for Photon Science and Synchrotron Radiation}\\
Eggenstein-Leopoldshafen, Germany \\
\textit{Laboratory for Applications of Synchrotron Radiation}\\
Karlsruhe, Germany \\
thomas.vandekamp@kit.edu}
\and
\IEEEauthorblockN{Andreas Kopmann}
\IEEEauthorblockA{\textit{Karlsruhe Institute of Technology (KIT)}\\
\textit{Institute for Data Processing and Electronics} \\
Eggenstein-Leopoldshafen, Germany \\
andreas.kopmann@kit.edu}
}




\IEEEoverridecommandlockouts
\IEEEpubid{\makebox[\columnwidth]{979-8-3503-2445-7/23/\$31.00~\copyright2023 IEEE \hfill}
\hspace{\columnsep}\makebox[\columnwidth]{ }}


\maketitle


\begin{abstract}

The unprecedented rate at which synchrotron radiation facilities are producing micro-computed (micro-CT) datasets has resulted in an overwhelming amount of data that scientists struggle to browse and interact with in real-time. Thousands of arthropods are scanned into micro-CT within the NOVA project, producing a large collection of gigabyte-sized datasets. In this work, we present methods to reduce the size of this data, scaling it from gigabytes to megabytes, enabling the micro-CT dataset to be delivered in real-time. In addition, arthropods can be identified by scientists even after implementing data reduction methodologies. Our initial step is to devise three distinct visual previews that comply with the best practices of data exploration. Subsequently, each visual preview warrants its own design consideration, thereby necessitating an individual data processing pipeline for each. We aim to present data reduction algorithms applied across the data processing pipelines. Particularly, we reduce size by using the multi-resolution slicemaps, the server-side rendering, and the histogram filtering approaches. In the evaluation, we examine the disparities of each method to identify the most favorable arrangement for our operation, which can then be adjusted for other experiments that have comparable necessities. Our demonstration proved that reducing the dataset size to the megabyte range is achievable without compromising the arthropod’s geometry information.  

\end{abstract}


\section{Introduction}

Synchrotron radiation micro-computed tomography (micro-CT) is an imaging technique that produces high-resolution three-dimensional (3D) images non-destructively. These images are composed of two-dimensional (2D) trans-axial projections of an object~\cite{chung2019overview, arhatari2023micro}. This enables the examination of intricate biological and synthetic materials with submicron resolution~\cite{bicer2016optimization}. At the KIT Imaging Cluster, the experiments of the NOVA project employ hard X-rays to examine arthropods and deposit them in an extensive storage system with datasets of several gigabytes in magnitude~\cite{van2013insect, schmelzle2017}. However, using traditional tools to navigate the vast data collection is laborious. Although a single dataset may not meet the big data criteria, the cumulative effect of thousands of datasets is overwhelming scientists’ capacity to browse and interact with them in real time. Given the unprecedented rate at which synchrotron radiation facilities produce data, the need for an efficient visual exploration system has become more pertinent and pivotal than ever before.


A tomographic workflow starts with the X-ray intensity projections of the samples being recorded continuously at various angles and hence produces a sequence of images in the sinogram domain. These images are then reconstructed into volumetric data by using algorithms such as filtered back projections~\cite{chung2019overview,brun2017syrmep}. In the post data acquisition stage, the final volumetric data will be used by scientists to perform further analysis. Particularly, the analysis of volumetric biological imaging data often requires isolating individual structures from the volumetric data by segmentation~\cite{van2011biological, van2018parasitoid, losel2020introducing}. This research represents the first comprehensive study to produce low-latency arthropod visual previews during data acquisition.




\begin{figure*}[t!]
  \centering
      \includegraphics[width=1.0\textwidth]{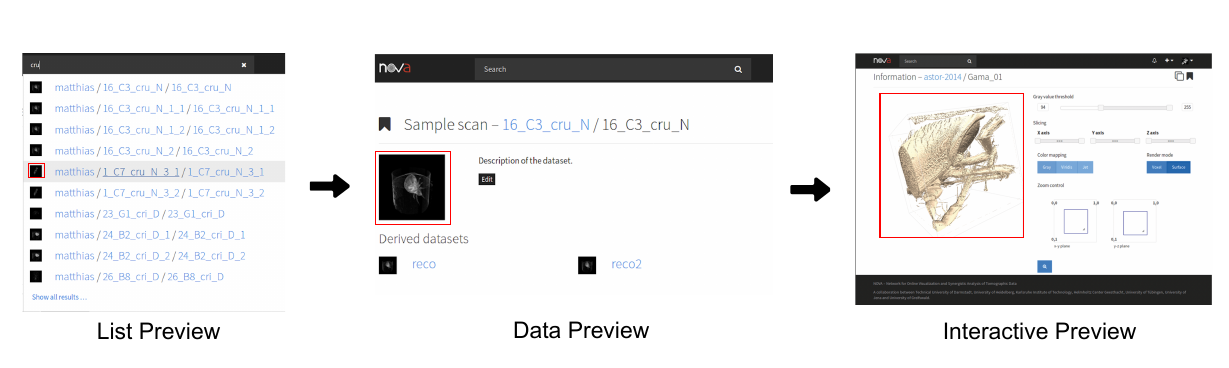}
  \caption{Classification of visual outputs based on the data exploration process which narrows down the data using top-down methodology, integrated within the NOVA data portal. The visual outputs in the three previews, namely list preview, data preview and interactive preview, are represented by the red boxes. Usually, users start by clicking on the dataset in the list preview. Subsequently, they are directed to the data preview page. If they wish to examine the dataset in a 3D view, they can simply click on the image within the data preview, which will then lead them to the interactive preview page.}
  \label{fig:classification_views}
\end{figure*}

\begin{table*}[htbp]
\caption{Overview of the data processing pipeline for the visual previews. The grey box represent the active process within each preview.}
\begin{center}
\begin{tabular}{|c|c|c|c|c|c|c|}
\hline
\textbf{Visual}&\textbf{Raw Data}& \textbf{3D} & \textbf{Slicemaps} &\textbf{Thresholding}&\textbf{Server-side}&\textbf{Histogram} \\
\textbf{Previews} & \textbf{(Slices)} & \textbf{Conversion}& \textbf{Conversion} & \textbf{+ Container Removal} & \textbf{Rendering} & \textbf{Filtering}\\
\hline
List Preview & \cellcolor{darkgrey} $\rightarrow$ & \cellcolor{darkgrey} $\rightarrow$ &  & \cellcolor{darkgrey} $\rightarrow$ & \cellcolor{darkgrey} $\rightarrow$ &  \\
\hline
Data Preview & \cellcolor{darkgrey} $\rightarrow$ & & \cellcolor{darkgrey} $\rightarrow$ & &  & \cellcolor{darkgrey} $\rightarrow$ \\
\hline
Interactive Preview & \cellcolor{darkgrey} $\rightarrow$ &  & \cellcolor{darkgrey} $\rightarrow$ & \cellcolor{darkgrey} $\rightarrow$ & & \\
\hline
\end{tabular}
\label{tab1}
\end{center}
\end{table*}

By generating visual previews of micro-CT datasets during data acquisition, scientists can quickly identify the type of arthropods visually without manual dataset labeling, thus facilitating the process of data identification. Therefore, an effective data browsing platform must resolve two primary challenges: perceptual scalability and data responsiveness. Perceptual scalability alludes to scientists’ ability to recognize the sample, post executing data reduction techniques. More precisely, the smaller datasets must conserve the geometric structure of the arthropods. To ensure data responsiveness, it is necessary that the datasets are available to users in real-time, regardless of their hardware requirements. Hence, we need to achieve a minimal latency between the server and the client. Two methods are available: first, the dataset can be rendered on the server-side, and then the resulting image can be transmitted to the client. Second, the server-side dataset can be reduced and then transmitted as a reduced volumetric dataset to the client~\cite{tanjerome2017wave, tan2019low}. The first method requires high hardware capability on the server-side, whereas the second approach would delegate the volume rendering responsibility to the client.

This paper will detail the implementation of data processing techniques that allow for a reduction in data size while retaining the geometrical structure of arthropods. The primary contributions we offer are:

\noindent (1.) Given the vast array of visual representations proposed in the literature, we first analyze the characteristics of visual outputs in a traditional data exploration system, leading us to identify three distinct visual outputs (Section 2).

\noindent (2.) We present data reduction methods that are based on the three visual outputs that can reduce the size from gigabytes to megabytes. A significant emphasis was placed on maintaining the geometry of the arthropods in the presented methods (Section 3). Notably, the ITS method, the optimal image approach and the histogram filtering introduce innovative approaches that are unique within this research domain. 

\noindent (3.) Our assessment of the different methods mentioned encompasses both visual and analytical approaches (Section 4).

\section{Design Considerations of Visual Previews}
\label{sec:visual_previews}

Using visual previews enables domain experts to narrow down and identify relevant data they are interested in. The preview terminology denotes a reduced version of the initial data that keeps its geometrical information~\cite{tanjerome2018digital}. To search a specific dataset, skilled data seekers were reported to follow the Visual Information Seeking Mantra, which recommends starting with an overview, then zooming in and filtering, and finally requesting details only when necessary. This data searching pattern is deemed to be the sure path to discovery.


Our adoption of this concept results in three distinct perspectives, with the list view showing an overview of all datasets, the data view detailing the chosen dataset, and the interactive view providing a comprehensive visualization of the dataset (refer to Figure~\ref{fig:classification_views}). The final visual outputs are denoted by the red boxes. Below, we outline each design consideration for every preview.

\textbf{List Preview}. The resultant visual output is a thumbnail image representing the dataset. Considering the arthropod’s three-dimensional nature, what is the most optimal method for generating a two-dimensional image that accurately represents the arthropod? What is the optimal viewing angle for creating an image snapshot? What is the metric that could automate the process of image generation? Through the answering of these questions, we shall produce a diminutive two-dimensional image that portrays the contour of the arthropods.

\textbf{Data Preview}. The data preview offers a thorough representation of the designated data, featuring an enlarged visual depiction. This preview should have more information than the list preview while having a rather small data size. One idea that could be interesting is to retain the outer geometry information while discarding the internal volume, resulting in a hollow dataset. Potentially, we could include object movement to augment object recognition.

\textbf{Interactive Preview}. The entire volumetric data will be loaded when using the interactive preview. In order to attain a visualization response in real-time, a preliminary dataset featuring a crude resolution is initially loaded, which is then followed by the primary dataset~\cite{tanjerome2017wave}. The characteristic enables users to choose a region of interest for further examination.

\section{Methods}

Within this section, we will explore approaches to reduce the primary dataset while complying with the design considerations stated in Section~\ref{sec:visual_previews}. An overview of the data processing methods employed to generate the final visual output of the previews is presented in Table~\ref{tab1}. The reconstructed micro-CT dataset constitutes the initial state, where 2D images are stacked together to create a volumetric series. Henceforth, we will refer to them as slices. Prior to starting the data processing pipeline for every visual preview, the slices are converted into a 3D object representation that will be subjected to data reduction operations. Within our context, the transformation of the slices is performed by converting them into either slicemaps~\cite{noguera2012visualization,tanjerome2017wave} or a 3D file format, such as OBJ-format. Subsequently, the data will undergo data reduction processes before being transferred to the final visual outputs.

\subsection{Thresholding}

The goal is to isolate specific regions of interest within the 3D volume, such as different tissue types, voids, or materials. In our context, we want to extract the arthropods from the surroundings. The process of thresholding aids in distinguishing between these regions through the application of binary classification to each voxel (3D pixel) based on its X-ray attenuation value. Tan Jerome et al. demonstrated the use of a real-time local noise filter and Otsu thresholding to reduce over-thresholding~\cite{tan2019real}. This filter is integrated into the GPU shader and will be used throughout our work. Therefore, we will use the Otsu threshold technique to determine the optimal threshold value. Additionally, a novel algorithm for threshold selection, named iterative threshold selection (ITS), has been formulated using a greedy algorithm approach. Below, we describe each of these algorithms.

\begin{algorithm}[h]
 $T_{otsu}$ $\leftarrow 0$, $\sigma_{otsu}$ $\leftarrow 0$\;
 \For{$T_{sweep}$ \text{from $0$...$255$} }{
     $R_a$ $\leftarrow$ histogram[$0$:$T_{sweep}$]\;
     $R_b$ $\leftarrow$ histogram[$T_{sweep}$:$255$]\;
     $W_a$ $\leftarrow$ density($R_a$), $W_b$ $\leftarrow$ density($R_b$)\;
     $U_a \leftarrow$ Mean($R_a$), $U_b \leftarrow$ Mean($R_b$)\;
     $\sigma \leftarrow W_a\times W_b\times (U_a-U_b)^2$\;
     
     \If{$\sigma > \sigma_{otsu}$}{
         $\sigma_{otsu} \leftarrow \sigma$\;
         $T_{otsu} \leftarrow T_{sweep}$\;
     }
 }
 \caption{Otsu thresholding}
 \label{alg:otsu}
\end{algorithm}

\textbf{Otsu thresholding}. The Otsu thresholding technique executes a conceptual sweep line to determine the most suitable threshold based on a criterion function that measures the statistical separation between the foreground and background classes. The criterion function entails the minimization of the ratio of between-classes variance and total variance (Equation~\ref{eq:between-class}).

\begin{equation}
\sigma^{2}_{\omega}(T) = \omega_{0}(T)\sigma^{2}_{0}(T) + \omega_{1}(T)\sigma^{2}_{1}(T),
\label{eq:between-class}
\end{equation}

\noindent where $\omega _{0}$ and $\omega _{1}$ are the weights which represent the probabilities of the two classes separated by the threshold $T$. Let $\sigma _{0}^{2}$ and $\sigma _{1}^{2}$ show the variances of the two types. The Otsu threshold is used to establish the lower limit of the intensity range (Algorithm~\ref{alg:otsu}). Starting from the initial grey value, $T_{otsu} = 0$, the histogram is partitioned into two regions, and the threshold that optimizes the between-class variance is chosen.

\textbf{Iterative threshold selection (ITS)}. The iterative threshold selection (ITS) method bears resemblance to greedy algorithms~\cite{edmonds1971matroids}. It aims to achieve a global minimum by computing the average intensities of the foreground and background clusters. The ITS method comprises five sequential stages, in which the second till fourth steps are reiterated until a threshold value converges (refer to Algorithm~\ref{alg:iterative} for details). In the first step, the algorithm selects a starting threshold, represented by $T_{it}$ from the interval of [0,255]. As the histogram distribution is concentrated in the middle of the dynamic range, the middle threshold can be considered a suitable initial point. Next, the histogram is partitioned into two regions, namely $R_1$ and $R_2$, determined by the chosen threshold $T_{it}$ (Step 2). In the third step, the mean intensity values for each region, denoted as $\mu_1$ and $\mu_2$, are computed. The fourth step requires an update to the threshold, which is determined by calculating the average of both mean intensities as $T_{it} = (\mu_1 + \mu_2) / 2$. Finally, Steps 2 through 4 are reiterated until the mean values $\mu_1$ and $\mu_2$ remain constant in consecutive iterations (Step 5). 

\begin{algorithm}[h]
 $T_{it}$ $\leftarrow$ $T_{start}$\;
 
 $r$ $\leftarrow$ $T_{start}$ \tcp*{$T_{it}$ converges if $r=0$}
 \While{$r \neq 0$}{
     $R_a$ $\leftarrow$ histogram[$0$:$T_{it}$]\;
     $R_b$ $\leftarrow$ histogram[$T_{it}$:$255$]\;
     $\mu_1$ $\leftarrow$ MeanIntensity($R_a$)\;
     $\mu_2$ $\leftarrow$ MeanIntensity($R_b$)\;
     
     $T_{tmp} \leftarrow (\mu_1 + \mu_2) / 2$\;
     $r \leftarrow  \lvert r - T_{tmp}\rvert$\;
     $T_{it}$ $\leftarrow$ $T_{tmp}$\;
 }
 \caption{Iterative threshold selection (ITS)}
 \label{alg:iterative}
\end{algorithm}

\subsection{Container Removal}

At the data acquisition stage, a cylindrical container made of 3D printing is used to house the biological samples. Therefore, the geometry of the container is scanned along with the collected data. An approach that can be feasible is to define the geometry of the container holding the sample and eliminate it from the data. Despite the uncomplicated cylindrical geometry of the sample container, the main obstacle lies in accurately determining the position and radius of said geometry. As a response, we make use of the Hough Circle Transform~\cite{ioannou1999circle} to recognize a circle from the top slice image. As the samples are at the base of the container, we assume that the aforementioned images delineate the geometry information. Lastly, we discard the information of the image that are outside of the determined circle, and apply this operation to every image slice.

\subsection{Server-side Rendering (Optimal Image Snapshot)}



Within this section, our proposed method will be presented to identify the most optimal viewpoint of a 3D dataset. This will be achieved by projecting the final 3D view space into a 2D image. The term "optimal" pertains to the perspective that encompasses the highest amount of information in a two-dimensional image. The method is advantageous because of the high-quality of the rendered image, which is directly derived from the raw 3D data.

In order to determine the optimal 2D image, we employ the Shannon entropy criterion~\cite{applebaum1996probability} as our comparison metric, given that the Shannon entropy indicates the image with the most information. The Shannon entropy, denoted as $H$, provides a statistical measure of randomness that is used to characterize the texture of the rendered image formally.

\begin{equation}
H = - \sum_{i=0}^{m-1} p_i \log _{2}\left(p_i\right),
\end{equation}

\noindent where $p_i$ contains the normalised histogram counts of the image. Greater entropy value implies increased information content in the generated image, enabling domain experts to identify the data more accurately.

\subsection{Histogram Filtering}

\begin{figure}[h!]
  \centering
      \includegraphics[width=0.45\textwidth]{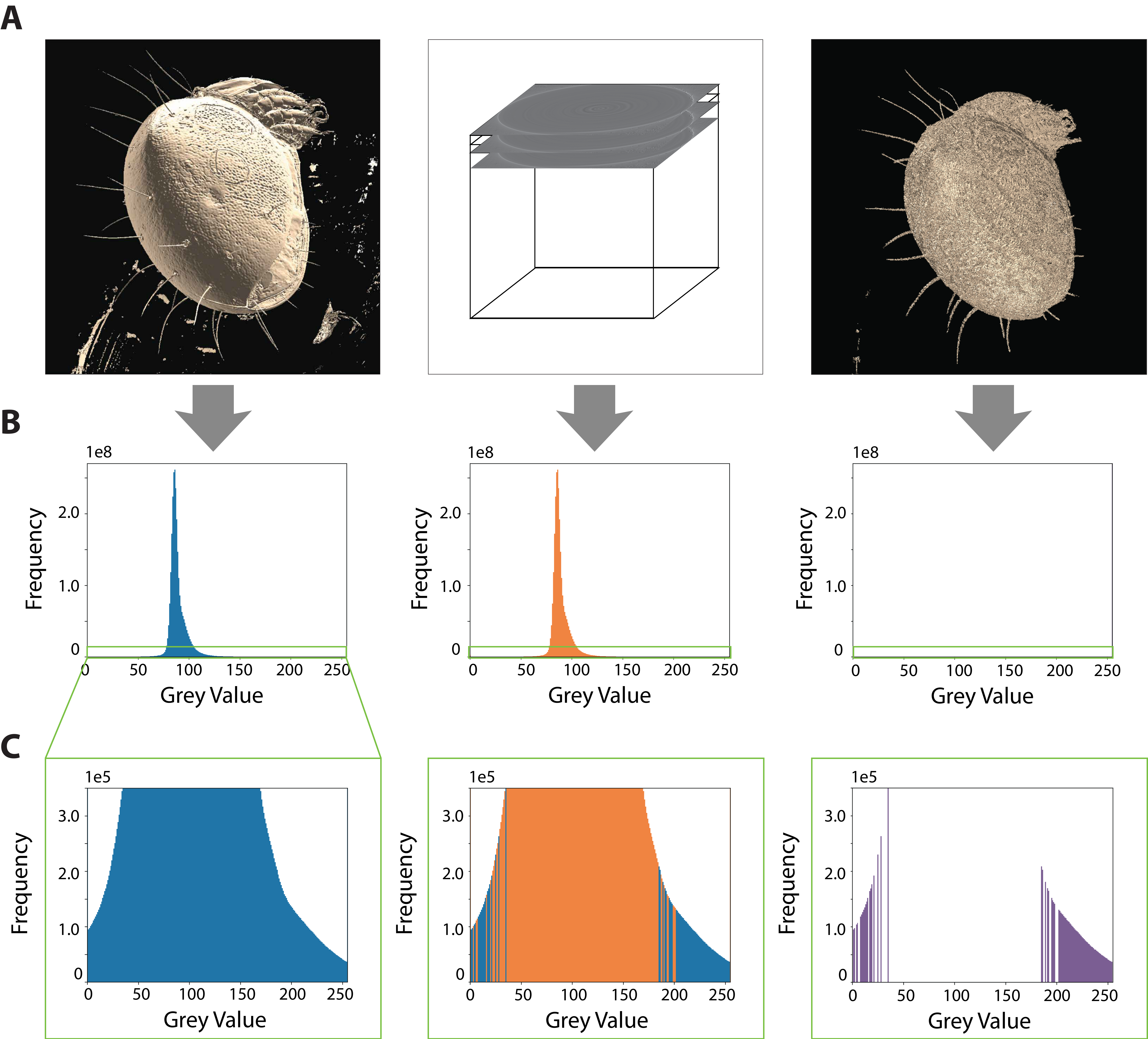}
  \caption{The data in its original form is presented in the left column, with the histogram distribution depicted in blue. Row A displays the surface rendering of the oribatid mite dataset (Table~\ref{tab:dl_datasets}: A), while row B shows the volumetric dataset's histogram. Row C, on the other hand, provides a zoomed-in view of the respective histogram distribution. The histogram distribution of the first three cross-section images (indicated in orange) that are artefacts is displayed in the middle column. The right column shows the resulting surface rendering of the filtered data (histogram distribution is shown in purple colour), where the unwanted bins are extracted from the original data.}
  \label{fig:noise_signature_removal}
\end{figure}

In order to streamline the data processing pipeline for the data preview, it is possible to decrease the dataset size by discarding the internal volume and retaining solely the external geometry data. The methodology postulates that the grey values of the surrounding artefacts, which may be air or the sample container, are represented by the top 3 slices of the volume. The validity of the assumption can be asserted since the object occupies only the lower portion of the sample container and does not fill up the entire space within it. From this point forward, we dispose the histogram bins containing undesirable grey values. The procedure of removing unwanted bins from the original data histogram is illustrated in Figure~\ref{fig:noise_signature_removal}. Consequently, only the sample characteristics were retained, albeit at the expense of finer details. The outcome indicated a surface that lacked intricate details and appeared rough.

\section{Evaluations and Discussions}

\begin{table}[t]
\caption{The information of datasets used in the evaluations.}
\begin{tabular}{ l l l l l l}
    \hline \\ [-1.5ex]
    Label & Type  & Image Resolution & Total Slices & Size\\[1.5ex]\hline
    \\ [-1.5ex]
    A & Oribatid mite &  $1536\times1536$ & 1152 & \SI{2.89}{GB}\\ [1.0ex] 
    B & Box mite & $1536\times1536$ & 1152 & \SI{2.72}{GB}\\ [1.0ex]
    C & Gammasid mite & $2016\times2016$ & 2016 & \SI{8.19}{GB}\\ [1.0ex]
    D & Pseudoscorpion & $2016\times2016$ & 1692 & \SI{6.88}{GB}\\  [1.0ex]
    E & Tachinid fly & $1968\times1968$ & 1456 & \SI{2.21}{GB}\\ [1.0ex]
     \hline
  \end{tabular}
  \label{tab:dl_datasets}
\end{table}

It can be observed from the three distinct visual previews that each data processing pipeline has its own design considerations, leading to differing approaches. Our attention is directed towards the two foremost challenges, namely perceptual scalability and data responsiveness. To address these challenges, we must decrease the dataset size to maintain data responsiveness while still achieving arthropod recognition. In this section, we will examine the visual outputs post the application of the data reduction method.

\subsection{Comparing Otsu and ITS Thresholding Approaches}

\begin{figure}[t!]
  \centering
      \includegraphics[width=0.40\textwidth]{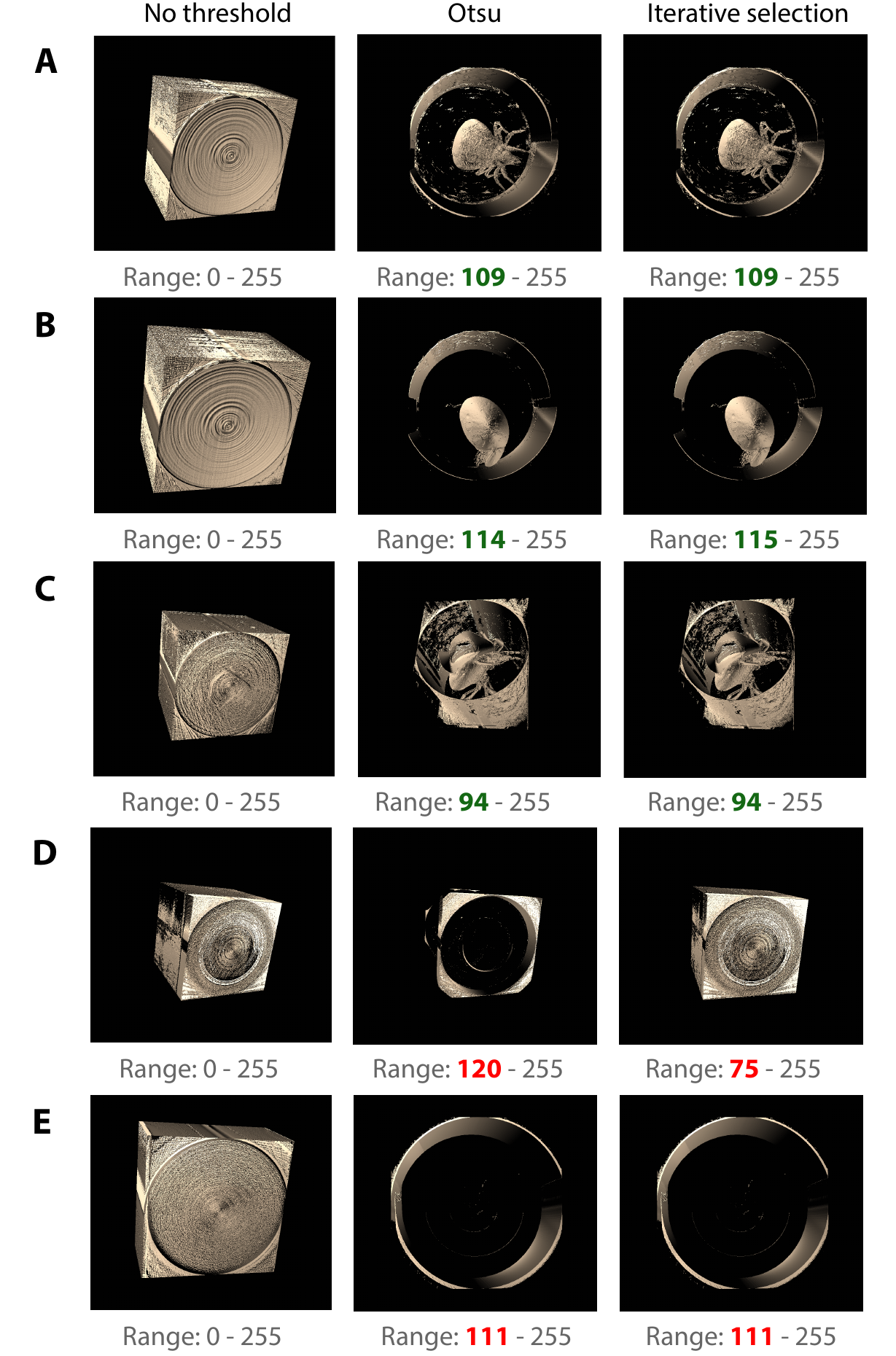}
  \caption{The visual results of datasets when applying threshold values calculated by the Otsu and iterative threshold selection (ITS) methods. The red colour shows the values that are under- or over-thresholded. The green colour shows the optimal threshold for each particular dataset.}
  \label{fig:volume_threshold}
\end{figure}

The fundamental variation between the two approaches resides in the number of iterations. Regarding Otsu thresholding, the algorithm consistently carries out a comprehensive scan throughout the entire dynamic range, from (\SIrange{0}{255}{} range) for an 8-bit data. On the contrary, the ITS method ceases its operation after discovering the global minimum, which frequently entails a significantly reduced number of iterations. The comparison of the two thresholding approaches has been conducted using the datasets described in Table~\ref{tab:dl_datasets}. The resulting visual outcomes are presented in Figure~\ref{fig:volume_threshold}. The Otsu and ITS methods are providing identical threshold values for the first three datasets (Table~\ref{tab:dl_datasets}: A, B, and C), thus emphasizing the shape of the samples. Despite this, the final two datasets (Table~\ref{tab:dl_datasets}: D and E) encountered difficulty in extracting biological specimens due to the restricted distribution of the histograms, resulting in the failure to obtain an optimal value. Given that Otsu and ITS methods rely on discrete threshold values, a minor shift within the narrow dynamic range produces a significant alteration in variance spreads (Otsu) or mean intensities (ITS). The container used in the sample was a contributing factor to the overall volume histogram, which, in turn, affected the analysis of the global histogram. Thus, the elimination of the sample container before the thresholding scan may result in an enhanced and precise analysis of the histogram. 

\begin{figure}[t!]
  \centering
      \includegraphics[width=0.40\textwidth]{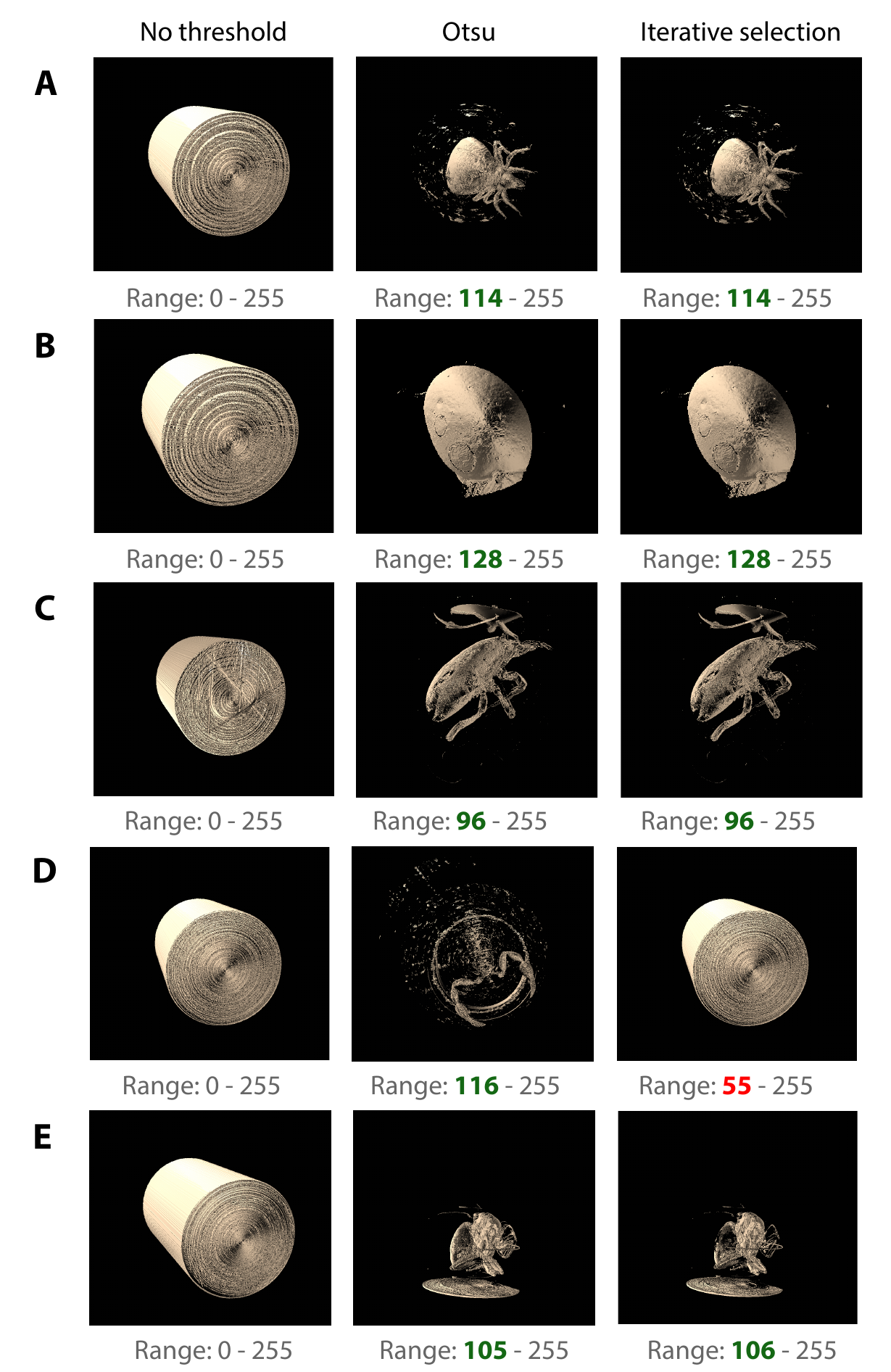}
      \caption{The visual results of cropped datasets when applying threshold values calculated by the Otsu and iterative threshold selection (ITS) methods. The red colour shows the values that are under- or over-thresholded. The green colour shows the optimal threshold for each particular dataset.}
  \label{fig:volume_threshold_crop}
\end{figure}

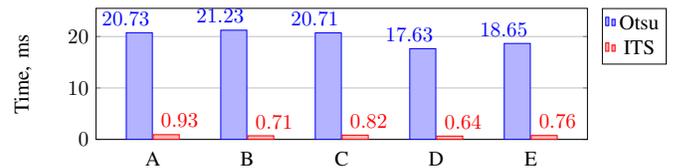
\begin{figure}[t!]
\centering
\resizebox{\linewidth}{!}{
\begin{tikzpicture}[thick,scale=1, every node/.style={scale=2}]
    \begin{axis}[
        width  = \textwidth,
        height = 6cm,
        major x tick style = transparent,
        ybar=2*\pgflinewidth,
        bar width=25pt,
        ymajorgrids = true,
        ylabel = {Time, ms},
        symbolic x coords={A,B,C,D,E},
        xtick = data,
        scaled y ticks = false,
        enlarge x limits=0.15,
        enlarge y limits={upper, value=0.2},
        ymin=0,
        nodes near coords,
        legend pos=outer north east
    ]
        \addplot
            coordinates {(A, 20.73) (B,21.23) (C, 20.71) (D, 17.63) (E, 18.65) };
            
        \addplot
            coordinates {(A, 0.9269) (B,0.7093) (C, 0.8231) (D, 0.6433) (E, 0.7626) };

        \legend{Otsu,ITS}
    \end{axis}
\end{tikzpicture}
}
\caption{The performances of the Otsu and the iterative threshold selection methods on a series of datasets described in Table~\ref{tab:dl_datasets}.}
\label{fig:otsu_its}
\end{figure}

The visual outcomes on the datasets after the execution of the container removal approach are showed in Figure~\ref{fig:volume_threshold_crop}. The shape of the samples in the datasets for the previous Pseudoscorpion and the tachinid fly (Table~\ref{tab:dl_datasets}: D and E) shows the significance of excluding the sample container. The Otsu and ITS methods were able to extract sample shapes more effectively after omitting sample container details from the histogram. However, the Pseudoscorpion dataset (Table~\ref{tab:dl_datasets}: D) remains a difficult task for the ITS method, while the Otsu approach exhibits robustness, even when confronted with a narrow histogram distribution. Considering the current dynamic range of the data, which is limited to 8-bit image, the Otsu thresholding method is deemed more reliable and thus, a superior thresholding method that should be implemented across all datasets. If datasets with higher dynamic range are available, the ITS method may be a more suitable alternative.

In order to attain a more comprehensive comparison of both algorithms on distinct datasets, we must initially convert the raw slices into slicemaps based on the $512^3$ scheme. The performance of each method is assessed by computing the average of ten computation runs on a MacbookPro having a \SI{64}{bit} Quad-Core Intel(R) Core i7 CPU at \SI{2.60}{GHz} and \SI{16}{GB} of DDR3 memory. Figure~\ref{fig:otsu_its} illustrates the average computation time for each dataset. The Otsu threshold exhibits a significantly prolonged computation time owing to the comprehensive scanning of the dynamic range of the image. The ITS approach relied exclusively on the convergence of the iterations and, using these datasets, the threshold converges after roughly ten iterations, enabling the ITS approach to execute 20 times faster than the Otsu method.

\begin{figure}[t!]
\centering
\resizebox{\linewidth}{!}{
\begin{tikzpicture}[thick,scale=1, every node/.style={scale=2}]
    \begin{axis}[
        width  = \textwidth,
        height = 6cm,
        major x tick style = transparent,
        ybar=2*\pgflinewidth,
        bar width=25pt,
        ymajorgrids = true,
        ylabel = {Time, s},
        symbolic x coords={A,B,C,D,E},
        xtick = data,
        scaled y ticks = false,
        enlarge x limits=0.15,
        enlarge y limits={upper, value=0.2},
        ymin=0,
        nodes near coords,
        legend pos=outer north east
    ]
        \addplot
            coordinates {(A, 2.10) (B, 3.28) (C, 3.78) (D, 3.40) (E, 3.49) };
           
    \end{axis}
\end{tikzpicture}
}
\caption{The performances of the finding the optimal 3D view point along Z-axis rotation using datasets described in Table~\ref{tab:dl_datasets}.}
\label{fig:optimal_3d}
\end{figure}
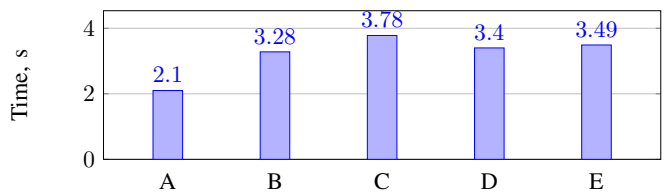


To evaluate the performance of the image snapshot approach, we chose five 3D datasets and performed Z-axis rotations to ascertain the highest entropy value in order to evaluate the system's effectiveness (Table~\ref{tab:dl_datasets}). It should be noted that the datasets are in 3D file format, which have been directly converted from the raw slices. The aggregate time taken for every process is documented in Figure~\ref{fig:optimal_3d}. Empirical evidence suggests that, on average, the system requires \SI{3.21}{\second} to identify the most informative 3D image via Z-axis rotation. The software for server-side rendering can be expanded to serve as a component for image streaming to distributed clients.

\section{Conclusion}
The unprecedented rate at which synchrotron radiation facilities are producing micro-CT datasets has resulted in an overwhelming amount of gigabyte-sized data that scientists struggle to browse and interact with in real-time. Within this work, we have established three distinct previews that conform to the best practices of data exploration. In the experiments of the NOVA project, arthropods are scanned into micro-CT resulting in thousands of datasets, with each in the gigabyte range in size. Our demonstration proved that reducing the dataset size to the megabyte range is achievable without compromising the arthropod's geometry information. In order to optimize data responsiveness, we have developed individual data processing pipelines for each aforementioned visual preview. The techniques we have presented are suitable for our particular use case and are modular, enabling customization for other experiments with comparable needs. Concerning future work, the methods displayed could be incorporated into a software framework to be helpful for the community. This has the potential to draw in more users and to subject the methods to further testing with additional micro-CT datasets.

\section*{Acknowledgment}

Data and/or analytical tools used in this study were provided by the projects ASTOR and NOVA (Michael Heethoff, TU Darmstadt; Vincent Heuveline, Heidelberg University; J{\"u}rgen Becker, Karlsruhe Institute of Technology), funded by the German Federal Ministry of Education and Research (BMBF; 05K2013, 05K2016).

\bibliographystyle{ieeetr}
\bibliography{ntj.bib}



\end{document}